\documentclass{article}

\PassOptionsToPackage{numbers, compress}{natbib}

\usepackage[preprint]{neurips_2020}

\usepackage[utf8]{inputenc} 
\usepackage[T1]{fontenc}    
\usepackage{hyperref}       
\usepackage{url}            
\usepackage{booktabs}       
\usepackage{amsfonts}       
\usepackage{nicefrac}       
\usepackage{microtype}      
\usepackage{graphicx}

\title{SpatialSim: Recognizing Spatial Configurations of Objects with Graph Neural Networks}

\author{%
  Laetitia Teodorescu\thanks{Use footnote for providing further information
    about author (webpage, alternative address)---\emph{not} for acknowledging
    funding agencies.} \\
  Inria\\
  Bordeaux, France \\
  \texttt{maria.teodorescu@inria.fr} \\
   \And
   Katja Hofmann \\
   Microsoft Research \\
   Cambridge, UK \\
   \texttt{katja.hofmann@microsoft.com} \\
   \And
   Pierre-Yves Oudeyer \\
   Inria \\
   Bordeaux, France \\
   \texttt{pierre-yves.oudeyer@inria.fr} \\
}

\begin{document}

\maketitle

\begin{abstract}
Recognizing precise geometrical configurations of groups of objects is a key capability of human spatial cognition, yet little studied in the deep learning literature so far. 
In particular, a fundamental problem is how a machine can learn and compare classes of geometric spatial configurations that are invariant to the point of view of an external observer.
In this paper we make two key contributions. First, we propose SpatialSim (Spatial Similarity), a novel geometrical reasoning benchmark, and argue that progress on this benchmark would pave the way towards a general solution to address this challenge in the real world. This benchmark is composed of two tasks: “Identification” and “Comparison”, each one instantiated in increasing levels of difficulty. Secondly, we study how relational inductive biases exhibited by fully-connected message-passing Graph Neural Networks (MPGNNs) are useful to solve those tasks, and show their advantages over less relational baselines such as Deep Sets and unstructured models such as Multi-Layer Perceptrons. Finally, we highlight the current limits of GNNs in these tasks.
\end{abstract}

\section{Introduction}

Our world appears to us as immediately organized into collections of objects, arranged together in natural scenes. These independent entities are the support for mental manipulation and language, and can be processed separately and in parallel \cite{pylyshyn2007things, 10.1093/bjps/axx055}. These objects are themselves composed of constituent elements whose precise arrangement in space determine their properties.

In contrast, recent successes in deep learning methods often rely on representations that do not exhibit such properties of organization. This means the compositional structure of the world has to be inferred from data, intuitively leading to slower, noisier learning. While such unstructured models have achieved important successes in a variety of domains, there is a strong emerging movement that advocates for the use of more structured models \cite{Lake_2016, DBLP:journals/corr/abs-1806-01261}, in particular for using models displaying relational inductive biases like Graph Neural Networks (GNNs). These are models that can approximate functions of graphs and that include in their architecture not only a shared feature space for all the nodes in the graph but also the ability to perform computation on the edges in the graph, e.g. based on the relationship exhibited by any two connected nodes. Various models based on these ideas have achieved substantial progress compared to unstructured methods in recent years, whether it is by considering the input as sets of objects in a shared space \cite{qi2016pointnet, zaheer2017deep} or by explicitly considering the relations between objects \cite{DBLP:journals/corr/BattagliaPLRK16}.

\begin{figure}[!t]
\vskip 0.2in
\begin{center}
\centerline{\includegraphics[width=330pt]{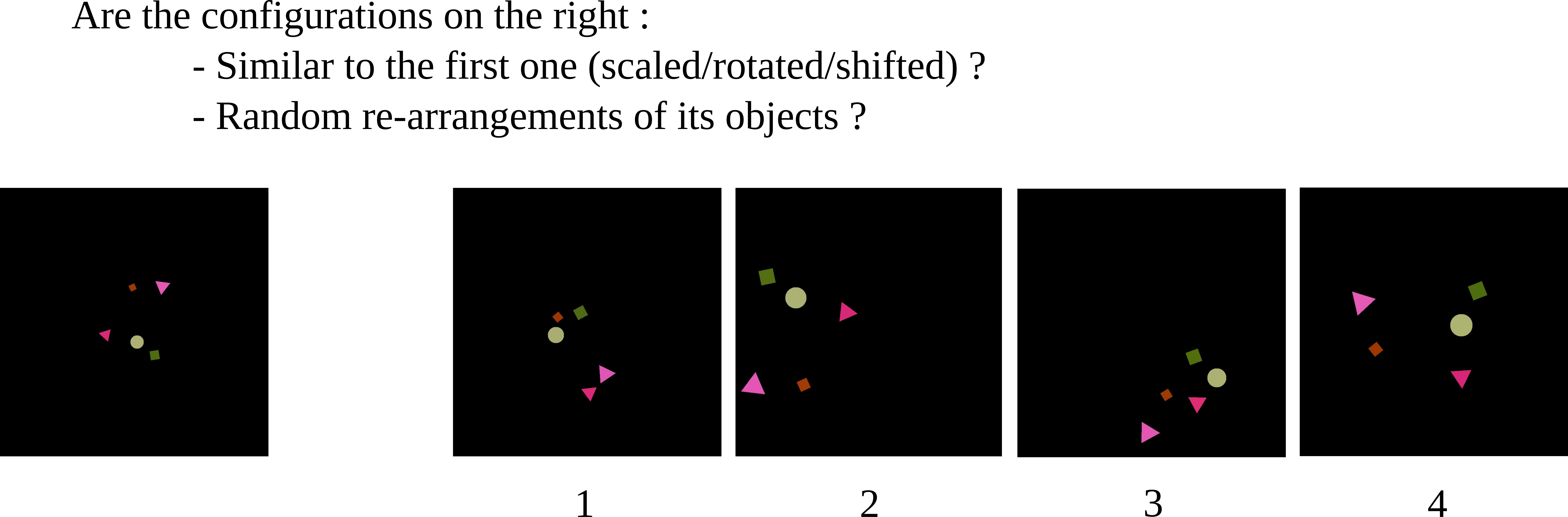}}
\caption{Visual illustration of SpatialSim. Which of the examples are positive samples, and which are negative? Answers in footnote \protect\footnotemark{}}
\label{visual-example}
\end{center}
\vskip -0.2in
\end{figure}

A particularly important set of tasks involving relational computation are those related to geometrical, or spatial, reasoning. We can describe scenes by naming their different elements and describe pairwise spatial relationships between them. But even before using words to describe, with limited precision, such relationships, we are able to precisely represent the relative positions occupied in space by different elements, allowing us for instance to form a representation of complex scenes that is invariant to our point of view. 
Such \emph{allocentric} spatial representations are a cornerstone of human intelligence \cite{ekstrom2017human}, heavily studied in child cognitive development \cite{vasilyeva2012development}, as well as a marker of cognitive aging \cite{moffat2006age}. 
For example, beyond the pervasive domain of spatial navigation \cite{ekstrom2017human}, there is considerable evidence that people process faces as precise configurations of distinct elements \cite{MAURER2002255}, where ratios of distances among elements influences the perception of naturalness. 
In the context of an agent embodied in the physical world, judging the stability of a construction of blocks \cite{DBLP:journals/corr/abs-1806-01203} requires appropriate summarizing of the geometrical relations between elements. In the context of an agent trying to reproduce a construction demonstrated by a teacher, the agent must be able to judge whether its current construction follows the same precise spatial configuration as the one demonstrated by the teacher, that may be presented from a different point of view. Indeed, when judging whether two configurations of objects are actually similar up to change of view of the observer, the fact that one object is on the right of another object provides no evidence towards the fact that the two configurations are the same or not, since this information can also be accounted by the change in perspective. This kind of spatial reasoning not only requires to perform relational computation on the objects but also to condition this computation on the global context. We thus see that reasoning on precise configurations of objects is a fundamental problem arising in classification, physical action, and imitation.

\footnotetext{Similar collections are Configurations 2 and 4}

To make progress towards solving the challenges motivated above we first need to be able to quantify progress. However, to our knowledge there is currently no controlled dataset or benchmark to tackle this problem directly. In this work, we introduce SpatialSim (Spatial Similarity), a novel spatial reasoning benchmark, to provide a well-defined and controlled evaluation of the abilities of models to successfully recognize and compare spatial configurations of objects. We divide this into two sub-tasks of increasing complexity: the first is called Identification, and requires to learn to identify a particular arrangement of objects; the second is called Comparison, and requires to learn to judge whether two presented configurations of objects are the same, up to a change in point of view. Furthermore, we test and analyse the performance of increasingly connected Message-Passing GNNs in this task. We find that GNNs that operate on fully-connected underlying graphs of the objects perform better compared to a less-connected counterpart we call Recurrent Deep Ret (RDS), to regular Deep Sets and to unstructured MLPs, suggesting that relational computation between objects is instrumental in solving the SpatialSim tasks. To summarize the key contributions of the paper:
\begin{itemize}
\item We introduce and motivate SpatialSim, a set of two spatial reasoning tasks and their associated datasets;
\item We compare and analyze the performance of state of the art GNN models on these two tasks. We demonstrate that relational computation is an important component of achieving good performance.
\item We provide preliminary analysis in the limits of these models in completely solving the benchmark.
\end{itemize}

\section{Related Work}

\subsection{Language-Conditioned spatial relations}

While previous work in Visual Question Answering (VQA \cite{VQA}) and Instruction-Following Reinforcement Learning \cite{bahdanau2018learning} have considered issues related to the ones we consider in SpatialSim, such work is constrained to using spatial relations that can be labelled by natural language. For instance, one of the standard benchmarks in VQA is the CLEVR dataset \cite{johnson2016clevr}. In CLEVR, questions are asked about an image containing a collection of shapes. The questions themselves are more designed for their compositionality ("the object that is of the same color as the big ball that is left of ...") than for geometric reasoning \emph{per se}. While the dataset does contain questions related to spatial reasoning, there are very few of them (four) and they are of a different nature than the precise reasoning on configurations that we wish to address. For instance, the concepts of "left of" or "right of", while defining broad spatial relations, are not invariant to the point of view of the observer. Their very "Question-Answering" nature makes those datasets unsuitable for investigating the question of learning to identify and compare precise geometrical arrangements of objects.

\subsection{Graph Neural Networks}

This is based on the recent line of work on neural networks that operate on graph-structured input. Seminal work \cite{1555942,4700287} involved updating the representations of nodes using a contraction map of their local neighborhood until convergence, a computationally expensive process. Follow-up work alleviated this, by proposing neural network models where each layer performs an update of node features based on the previous features and the graph's adjacency matrix, and several such layers are stacked to produce the final output. Notably, Graph Convolutional Networks (GCNs) \cite{DBLP:journals/corr/KipfW16} \cite{DBLP:journals/corr/DefferrardBV16} \cite{ae482107de73461787258f805cf8f4ed}, in their layers, use a linear, first-order approximation of spectral graph convolution and proved effective in several domains \cite{NIPS2015_5954}. However, these works have focused on working on large graphs where the prediction at hand depends on its connectivity structure, and the GCN can learn to encode the structure of their k-hop neighborhoods (for k computation steps). In our case there are a lot less objects, and the precise connectivity is irrelevant since we consider GNNs on fully-connected graphs.

A different class of networks (MPGNNs) has been proposed to more explicitly model an algorithm of message-passing between the nodes of a graph, using the edges of the graph as the underlying structure on which to propagate information. This is the class of model that we consider in this work, because of their flexibility and generality. A variant of this was first proposed to model objects and their interactions \cite{DBLP:journals/corr/BattagliaPLRK16,NIPS2017_7082} by explicitly performing neural computation on pairs of objects. The full message-passing framework was introduced in \cite{DBLP:journals/corr/GilmerSRVD17} for supervised learning on a molecular dataset, and was expanded in \cite{DBLP:journals/corr/abs-1806-01261}, which provided a unifying view of existing approaches.

A line of theoretical work has focused on giving guarantees on the representational power of GNNs. \cite{DBLP:journals/corr/abs-1810-00826} introduced a simple model that is provably more powerful than GCNs, and universality of approximation theorems for permutation invariant \cite{DBLP:journals/corr/abs-1901-09342} and permutation equivariant \cite{DBLP:journals/corr/abs-1905-04943} functions of nodes of a graph have also derived recently. However, the models constructed in these proofs are theoretical and 
cannot be tractably implemented, so this line of work opens interesting avenues for empirical research in the capacities of GNNs in different practical settings. In particular, \cite{DBLP:journals/corr/abs-1810-00826} found that GNNs have powerful theoretical properties if the message-aggregation function is injective. This means that the aggregation function has to preserve all the information from its input, which is a collection (of a priori unbounded size) of vectors, to its output, which is a single vector of bounded dimension. Our setting allows us to examine the ability of GNNs to appropriately aggregate information in a practical, naturally-occurring setting.

Finally, our work relates to recent work on learning distance functions over pairs of graphs \cite{ma2019deep}. Recent work \cite{DBLP:journals/corr/abs-1904-12787}, in a model they call Graph Matching Network, has proposed using a cross-graph node-to-node attention approach for solving the problem, and compared it to an approach without cross-graph attention, that is closely related to our models for the Comparison task. However, our work here is distinct, not so much in the actual architectures used, but in the properties of the setting. Their work considered a task related to graph isomorphism, where the precise connectivity of the graph is important. In this work we are interested in the features of the nodes themselves and we consider the underlying graph simply as a structure supporting the GNN computation.

\section{The SpatialSim benchmark}

\subsection{Description}

In this work, we consider the problem of learning to recognize whether one spatial configuration of objects is equivalent to another. The notion of equivalence that we consider is grounded in the motivation outlined above: train models that can reason on configurations of objects regardless of their point of view. Because of that, we define equivalence in SpatialSim as geometrical similarity, e.g. any arbitrary composition of translations, rotations, and scalings. In general terms, we frame the problem as a classification problem, where positive examples are drawn from the same similarity equivalence class, and negative examples are drawn from a different one. Since we want, for this first study, to provide the simplest possible version of the problem, we place ourselves in the 2d plane. Extending the setting to 3 dimensions does bring some additional complexity (rotations are also described by a direction and not only by an angle and a center), but does not change the underlying mathematical problem. We provide a visual illustration of the setting in Figure \ref{visual-example}.


To provide a clean and controlled benchmark to study the ability of models to learn spatial similarity functions, we work with object features that are already given to the model. We ask the question: supposing we have a perfect feature extractor for objects in a scene, are we able to learn to reliably recognize spatial configurations? Working with objects directly allows us to disentangle the feature extraction process from the spatial reasoning itself. To further investigate the role of recent unsupervised object-extraction techniques \cite{burgess2019monet,greff2019multiobject,engelcke2019genesis} and assess the quality of the representations learned by such models in the context of SpatialSim we also provide corresponding image datasets, and leave experiments on these images to further work.

We define a set of $n_{obj}$ objects as colored shapes in 2d space, each uniquely characterized by a 10-dimensional feature vector. There are three possible shape categories, corresponding to squares, circles and triangles; the shape of each object is encoded as a one-hot vector. The shapes are distributed in continuous 2d space, and their colors belong to the RGB color space, where colors are encoded as a floating-point number between 0 and 1. The objects additionally have a size and an orientation, the latter given in radians. The feature vector for each object is the concatenation of all the previously described features, and a scene containing several objects is given as a set of the individual feature vectors describing each object. Note that the objects are unordered, and any permutation of the objects is a possible encoding for a particular scene.


We subdivide SpatialSim in two tasks. The first one, Identification, allows us to evaluate the abilities of different models to accurately summarize all relevant information to correctly respond to the classification problem. In this task, we sample a random configuration that will be the one the model in required to learn to identify. One configuration corresponds to one datasets, and we evaluate the capacities of the models on a set of configurations. The second task, Comparison, allows us, in addition to the study allowed by the Identification task, to judge whether the computation learned by a model can be trained to be universal across configurations. For this purpose, the task requires to predict whether two distinct presented configurations belong to the same class or not. We give additional details on data generation for those two settings in the following sections, and a summary table in the annex.

\subsection{First task: Identification}

The Identification task is composed of several reference configurations of $n_{obj}$ objects, each corresponding to a distinct dataset. Each sample of one dataset is either in the same similarity equivalence class as the reference configuration, in which case it is a positive example; or in a different similarity class as the reference, in which case it is a negative example. The same objects are present in all samples, not to give the model any additional information unrelated to spatial configurations.

This simple task allows us to isolate how well the model is able to learn as a function of the number of objects $n_{obj}$: indeed, the model must make a decision that depends on the relationship between each objects, and for this purpose has to aggregate incoming information from $n_{obj}$ vectors: when $n_{obj}$ gets large the information the model is required to summarize increases. This can lead to loss of performance is the capacity of the models stays constant. For this reason, we structure the task with an increasing number of objects: we generate 27 datasets, with $n_{obj} \in [3..30]$. We further group them into 3 collections of low number ($n_{obj} \in [3..8]$), medium number ($n_{obj} \in [9..20]$) and high number ($n_{obj} \in [21..30]$). 

For a given number of objects $n_{obj}$ and based on one reference configuration generated by sampling $n_{obj}$ random objects uniformly in 2d space, we generate a balanced dataset composed of:

\textbf{Positive examples}: after applying a small perturbation of factor $\varepsilon$, to each of the objects in the reference, very slightly changing their color, position, size and orientation, we apply a rotation around the configuration barycenter $B$ with an angle $\phi \sim \mathcal{U}([0, 2\pi])$, a scaling,of center $B$ and magnitude $s \sim \mathcal{U}([0.5, 2])$, and a translation of vector $t$. The latter is sampled from the same distribution as the positions of the different objects;

\textbf{Negative examples}: these examples are generated by applying a small perturbation to the features of the objects in the reference, slightly changing their colors and sizes, and then re-sampling randomly the positions of the objects, while keeping the object's identity (shape, size, color). After randomly resampling the objects' positions, we apply a rotation, scaling and translation drawn from the same distribution as in the positive example to all objects. This is done to ensure no spurious correlations exist that could help models identify positive from negative examples regardless of actual information about the configuration. While it is, in theory, possible to sample a negative example that is close to the positive class, the probability is very low in practice for numbers of objects $n \geq 3$.

For each reference configuration, we generate 10,000 samples for the training dataset, and 5,000 samples, from the same distribution, for the validation and test datasets. For each dataset, we train a model on the train set and test it on the test set. The obtained test accuracies of the models over all datasets are averaged over a group (low, medium and high number of objects).

\subsection{Second task : Comparison}

In this section, we describe our second task. While in the previous setup the model had to learn to identify a precise configuration, and could learn to perform computation that does not generalize across configurations, we envision Comparison as a more complex and complete setting where the model has to learn to \emph{compare} two different configurations that are re-drawn for each sample. This task, while being more difficult than Identification, is also more general and more realistic : while sometimes an agent may be confronted with numerous repetitions of the same configuration that it has to learn to recognize - for instance, humans become, by extensive exposition, quite proficient at the task of recognizing the special configuration of visual elements that is human faces - but a very common task an intelligent agent will be confronted to is entering a new room filled with objects it knows but that are arranged in a novel way, and having to reason on this precise configuration.

For this task, because each sample presents a different set of objects, $n_{obj}$ can vary from one sample to the other, and thus a single dataset can cover a range of number of objects. We generate three distinct datasets, one with $n_{obj} \in [3..8]$, one with $n_{obj} \in [9..20]$ and one with $n_{obj} \in [21..30]$. In preliminary experiments we have observed learning the Comparison task is very hard, leading to a great dependence on the initialization of networks: some seeds converge to a good accuracy, some don't perform above chance. This is due to the presence of rotations in the allowed transformations in the positive examples; a dataset containing only translations and scalings leads to good learning across initializations. Note that this problem with rotations persists for a simplified setup containing only configurations of $n_{obj} = 3$ objects. To alleviate this and carry the optimization process we introduce a curriculum of five datasets, each one with a different range of allowed rotations in the generation process, with the last one spanning all possible rotation angles.

We generate the dataset as:

\textbf{Positive examples}: we draw the first configuration by randomly sampling the objects' shape, size, color, position and orientation. For obtaining the second configuration, we copy the first one, apply a small perturbation to the features of each object, and apply a random rotation, scaling, and translation to all objects, using the same process as described in the Identification task;

\textbf{Negative examples}: we draw the first configuration as above, apply a small perturbation of magnitude $\varepsilon$ and for the second one we randomly re-sample the positions of each object independently, while keeping the other features constant. We finally apply a random rotation, scaling and translation and this gives us our second set of objects.

For each range of number of objects and for each dataset of the curriculum we generate 100,000 samples for the training set. We generate a validation and a testing set of 10,000 samples for each range of $n_{obj}$. Those datasets contain rotations in the full range.

\section{Models and Architectures}

With the benchmark, we establish a first set of reference results from existing models in the in the literature, serving to identify their strengths and limits, and as baselines for further work. We consider Message-Passing GNNs for their established performance, notably in physical reasoning tasks, along with stripped-down versions of the same models. Additionally, our hypothesis is that models that implement relational computation between objects will perform best in this benchmark, because it requires taking into account the relative positions between objects and not only their absolute positions in 2d-space. To test this hypothesis, we model a configuration of objects as a graph, where the individual objects are the nodes. We then train three GNN models with decreasing levels of intra-node communication: MPGNN (full Message-Passing GNN \cite{DBLP:journals/corr/GilmerSRVD17}) performs message-passing updates over the complete graph where all object-object edges are considered; RDS (Recurrent Deep Set) is a Deep Set model \cite{zaheer2017deep} where each object updates its features based on its own features and a global vector aggregating all the other object features (all-to-one message passing); and a regular Deep Set model where each node updates its own features independently.

We additionally compare, for both tasks, our models to Multi-Layer Perceptron (MLP) baselines. Our MLP baselines are built to have the same order of number of hidden units as the GNNs, to allow for similar representational capacity. However, because there is a considerable amount of weight sharing in GNNs compared to MLPs the number of weights is much higher, and additionally, increases substantially with the number of objects. More details on the MLP baselines are given in the supplementary material.

For all GNN models, after $N$ node updates, the node features are summed, passed through an MLP to output a two-dimensional vector representing the score for the positive and negative class. For the Comparison task, the models take as input two configurations, pass them through two parallel GNN layers, concatenate their output embeddings and pass this vector through a final MLP to produce scores for both classes. For a detailed description of all models, including a visual overview, please refer to the annex.

\section{Experimental Results}

\subsection{Identification}

In this section we report the experimental results for the Identification task. Reported results are averaged over 10 independent runs and across datasets in the same group (low, medium, high number of objects). For a fair comparison across parameter numbers in the GNNS, we build each internal MLP used in the message computation, node-wise aggregation, graph-wise aggregation and prediction steps with the same number of hidden layers $d$ and the same number of hidden units $h$. Since the DS and RDS layer can be seen as stripped-down versions of the MPGNN layer, if $d$ and $h$ stay constant the number of parameters drops for the RDS layer, and drops even further for the DS layer. To allow for a fair comparison with roughly the same number of parameters in each model, we use $h = 16$ for all architectures and $d = 1$ for the MPGNN, $d = 2$ for the RDS and $d = 4$ for the DS, and we report the number of parameters in the table. For the embedding vector that aggregates the whole configuration information, we use a dimension $h_{u} = 16$: thus the number of objects is at first small compared to $h_{u}$ and gradually becomes larger as $n_{obj}$ increases. We found no significant difference between using one or several rounds of node updating in the GNNs, so all our results were obtained with $N = 1$. We train for 20 epochs with the Adam optimizer \cite{kingma2014adam}, with a learning rate of $10^{-3}$.

We report the means and standard deviations of the test accuracies across all independent runs in Table \ref{results-simple}, as well as the numbers of parameters for each model. Chance performs at 0.5. We observe the highest accuracy with the MPGNN model, on all three object ranges considered. It achieves upwards of 0.97 percent accuracy, effectively solving the task for numbers of objects ranging from 3 to 30. Note that in this range performance of the MPGNN stays constant when the number of objects increases. In contrast, the RDS model achieves good performance (0.91) when the number of objects is low, but its performance decreases as the number of objects grows. Deep Sets show lower performance in all cases, and the MLP achieves 0.85 mean accuracy with $n_{obj} \in [3..8]$ but its performance drops sharply as $n_{obj}$ increases.

Our results provide evidence for the fact that, while it is possible to identify a given configuration with well-above chance accuracy without performing any relational computation between objects, to effectively solve the task it seems necessary to perform fully-connected message-passing between the objects, especially when the number of objects increases. However, while the number of parameters stays constant in a fully-connected MPGNN when $n_{obj}$ increases, the amount of computation scales as the number of edges ($\mathcal{O}(n_{obj}^2$)). This makes using MPGNNs on complete graphs harder to use at scale. However, note that we consider this task as an abstraction for naturally-occurring configurations of objects that contain a limited number of objects, so this quadratic increase in time complexity should not be a problem in practice.

\begin{table}[!t]
\caption{Test classification accuracies (means and standard deviations are given over datasets and seeds) for the three different models on the Identification task.}
\label{results-simple}
\vskip 0.15in
\begin{center}
\begin{small}
\begin{sc}
\begin{tabular}{lcccr}
\toprule
Model & $n_{obj} \in [3..8]$ & $n_{obj} \in [9..20]$ & $n_{obj} \in [21..30]$ & Parameters \\
\midrule
MPGNN        & \textbf{0.97}$\pm$ 0.026 & \textbf{0.98} $\pm$ 0.024 & \textbf{0.98} $\pm$ 0.028 & 2208\\
RDS          & 0.91 $\pm$ 0.062 & 0.85 $\pm$ 0.128& 0.78 $\pm$ 0.19 & 2038\\
Deep Set     & 0.65 $\pm$ 0.079 & 0.60 $\pm$ 0.082 & 0.58 $\pm$ 0.09 & 2386 \\
MLP Baseline & 0.82 $\pm$ 0.09 & 0.59 $\pm$ 0.051 & 0.56 $\pm$ 0.051 & 6k/48k/139k \\
\bottomrule
\end{tabular}
\end{sc}
\end{small}
\end{center}
\vskip -0.1in
\end{table}

\subsection{Comparison}

\begin{table}[!b]
\caption{Test classification accuracies for the three different models on the Comparison task. All metrics were computed on 10 different seeds and trained for 5 epochs on each dataset of the curriculum.}
\label{results-double}
\vskip 0.15in
\begin{center}
\begin{small}
\begin{sc}
\begin{tabular}{lcccr}
\toprule
Layer Type    & $n_{obj} \in [3..8]$ & $n_{obj} \in [9..20]$ & $n_{obj} \in [21..30]$ & Parameters \\
\midrule
MPGNN         & \textbf{0.89} $\pm$ 0.030 & \textbf{0.81} $\pm$ 0.121 & \textbf{0.71} $\pm$ 0.176 & 4686 \\
RDS           & 0.8 $\pm$ 0.133  & 0.68 $\pm$ 0.154 & 0.52 $\pm$ 0.04 & 5326 \\
Deep Set      & 0.51 $\pm$ 0.014  & 0.50 $\pm$ 0.001 & 0.50 $\pm$ 0.005  & 5274 \\
MLP Baseline  & 0.55 $\pm$ 0.002 & 0.51 $\pm$ 0.006 & 0.50 $\pm$ 0.004 & 26k/192k/552k  \\
\bottomrule
\end{tabular}
\end{sc}
\end{small}
\end{center}
\vskip -0.1in
\end{table}

We now turn to our Comparison task. We compare the dual-input architecture with different internal layers: MPGNN, RDS and DS, and to an MLP baseline. The MLP is built according to the principle stated above, and takes as input a concatenation of all the objects in both configurations.  As in the previous section, to ensure a fair comparison between models in terms of the number of parameters we provide our RDS and DS layers with deeper MLPs (see section on model details in supplementary material). We train models on three datasets, respectively with $n_{obj} \in [3..8]$, $n_{obj} \in [9..20]$ and $n_{obj} \in [21..30]$, for 10 seeds per dataset, in Table \ref{results-double}.

We report mean accuracies of the different models and their standard deviation in Table \ref{results-double}. As before, chance performance is 0.5. We immediately see the increased difficulty of Comparison compared to Identification: the model based on MPGNN layers performs best, with mean accuracies of 0.89, 0.81 and 0.71, compared to a performance above 0.97 on Identification. In this case we see the performance drop for MPGNN when $n_{obj}$ increases. RDS performs well above chance when the number of objects is small, but its performance drops rapidly afterwards. Both Deeps Sets and the MLP fail to reliably perform above chance. We established in the previous section that complete message-passing between nodes is instrumental in learning to identify particular configurations. This experiment suggests that layers that allow nodes to have access to information about other nodes are key to achieve good performance, whether this communication is centralized, via conditioning by the graph-level feature as in the RDS, or decentralized, as allowed by the MPGNN layer. Additionnaly, full node-to-node communication seems to be crucial for good performance, and we show in the next subsection how this affects the functions learned by the models. However, it does not seem to be enough to completely solve the task. We provide additional details on the generalization properties of the models in the supplementary material.

\subsection{Model Heat maps}

\begin{figure}[!t]
\vskip 0.2in
\begin{center}
\centerline{\includegraphics[width=460pt]{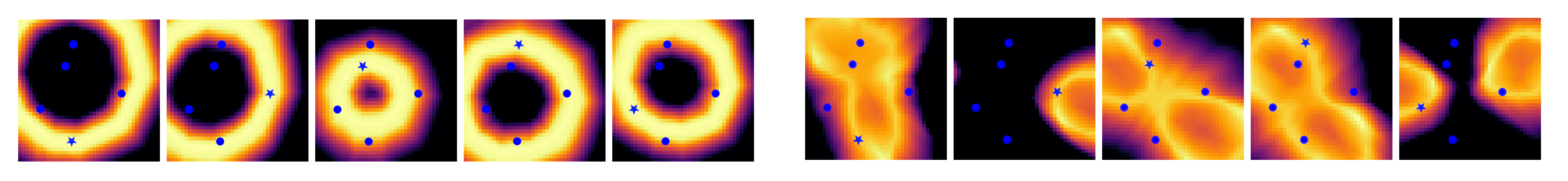}}
\caption{Magnitude of the difference in predicted score for the positive and negative classes for a comparison between a 5-object configuration and a perturbed version of this configuration where one object is displaced over the 2d-plane. Value displayed is $score_{+} - score_{-}$.Left row is with RDS layers and right row is with MPGNN. For each row, the displaced object's position is indicated with a blue star, the other ones with a blue dot. The sizes, colors, orientations and shapes of the objects are not represented. Bright yellow means the model assigns the positive class to the configuration where the displaced object would be placed here, black means the negative class would be assigned. Best viewed in color.}
\label{heatmaps}
\end{center}
\vskip -0.2in
\end{figure}

In addition to those quantitative results, we assess the quality of the learned functions for different models. We do this by visualizing the magnitude of the difference between the scores of the positive and negative classes, as output by the different models, as a function of the position of one of the objects in the configuration. We show the results in \ref{heatmaps}. Beyond the clear qualitative differences between different models, this figure clearly shows the shortcomings of the models. A perfect model for this task would show a high-magnitude region in the vicinity of the considered object, and low values everywhere else. Ideally, the object would be placed at a global maximum of this score difference function. Instead, both the RDS and MPGNN models show extended crests of high magnitudes: this means that moving the considered object along those crests would not change the prediction of the models, whereas the configurations are clearly changed. This suggests that the models we considered are limited in their capacity to distinguish classes of similar configurations. This lack of a clearly identifiable global maximum over variations of the position of one object suggests a possible reason for ceiling in performance exhibited by our models: the tested GNNs are unable to break certain symmetries. For the RDS, since each node only has access to global information about an aggregate of the other nodes, it is not surprising to see it exhibit the radial symmetry around the barycenter of the configuration. MPGNNs seem to operate in a different way: for each node the learned function seems to show symmetry around axes related to the principal directions of the distribution of other nodes. The models are thus insensitive to a great range of specific perturbations of the individual objects (eg the RDS is insensitive to the rotation of one object around the barycenter of the configuration, since the object stays in the high-magnitude zone). More discussion on this subject can be found in the appendix.

\section{Conclusion}

In this work, we motivated and introduced SpatialSim, a simplified but challenging spatial reasoning benchmark that serves as a first step towards more general geometrical reasoning where a model has to learn to recognize an arrangement of objects irrespective of its point of view. We demonstrated that the relational inductive biases exhibited by Message-Passing GNNs is crucial in achieving good performance on the task, compared to a centralized message-passing scheme or to independent updating of the objects. MPGNNs achieve near-perfect performance on the configuration identification task, but achieve much lower performance on the configuration comparison task. Our analysis suggests two shortcomings of current models on this benchmark: 1) the models struggle to accurately summarize information when the ratio between the number of objects and the size of the embedding used for representing the whole configuration becomes large; and 2) GNNs in spite of their relational inductive biases struggle to break certain symmetries; we take this to mean additional theoretical and experimental research is warranted to find more appropriate biases for geometrical reasoning.



\section*{Links}

The code, datasets and dataset generators are available at the following address: \href{https://sites.google.com/view/gnn-spatial-reco/}{https://sites.google.com/view/gnn-spatial-reco/}.

\section*{Broader Impact}

We present a new geometrical reasoning problem for the machine learning community, along with an associated benchmark and preliminary results. Our work thus contributes to building models with improved spatial reasoning capacities. Furthermore, by releasing our code and datasets, we hope to foster new research on geometrical reasoning abilities of machine learning models.

We believe that this work is too fundamental to directly cause significant ethical and societal impact. Indirectly, this work may open additional research on the spatial reasoning abilities of robots, leading for instance to improved imitation and navigation abilities. In addition to that, we hope this work can serve as the basis for further investigation and better understanding of the properties of AI methods, providing better opportunities to limit their potential negative effects and harness their potential positive effects on society.

\bibliography{example_paper}
\bibliographystyle{icml2019}

\end{document}




\section*{Supplementary Material: Summary}

This document provides additional details on the SpatialSim benchmark, the architectures and models used, and some additional experimental results and analysis. It is organized in the following way:


\begin{itemize}
    \item \textbf{Section \ref{section:benchmark}: SpatialSim Benchmark Summary}
    \item \textbf{Section \ref{section:dscreation}: Additional Details on Dataset Generation}
    \item \textbf{Section \ref{section:models}: Models and Architectures}
    \item \textbf{Section \ref{section:mhms}: Model Heatmaps; Additional Discussion}
    \item \textbf{Section \ref{section:easyhard}: Easier and Harder Configurations}
    \item \textbf{Section \ref{section:gen}: Generalization over Object Number}
    \item \textbf{Section \ref{section:trainvar}: Effects of Variations in Number of Training Examples}
    \item \textbf{Section \ref{section:distractors}: Adding Distractor Objects}
\end{itemize}

\section{SpatialSim Benchmark Summary} \label{section:benchmark}

This section provides a summary of the SpatialSim benchmark. 

The datasets, as well as the code and instructions to reproduce our experiments, are accessible at the following link: \href{https://sites.google.com/view/gnn-spatial-reco/}{https://sites.google.com/view/gnn-spatial-reco/}. 
We also provide the dataset generation code to produce extended versions of our datasets.

\begin{center}
\begin{figure}[!h]
    \centering
    \includegraphics[width=420pt]{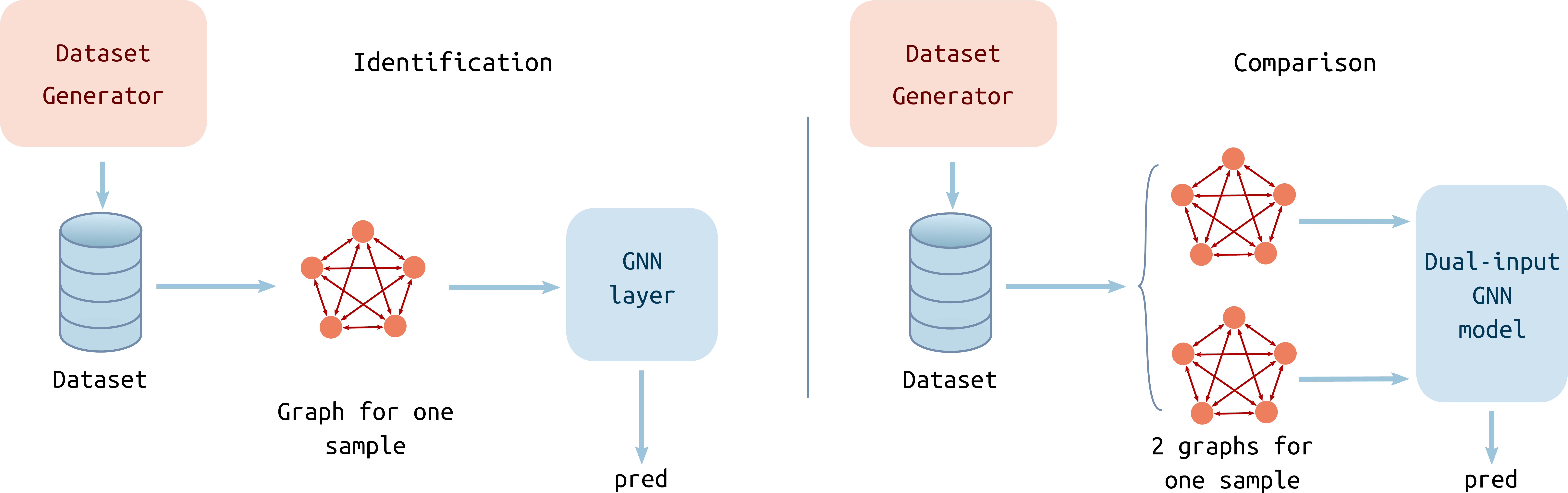}
    \caption{Schematic of the benchmark: we generate datasets for Identification and Comparison tasks. Each Identification data sample is transformed into an input scene graph. Each Comparison data sample is transformed in two input scene graphs.}
    \label{fig:summary_figure}
\end{figure}
\end{center}

All datasets belonging to both SpatialSim tasks are detailed in Table \ref{summary_table}. A visual illustration of the benchmark is given in Figure \ref{fig:summary_figure}.

\begin{table}[!h]
\begin{center}
\begin{tabular}{ | m{2.3cm} || m{2.3cm}| m{2.3cm} | m{2.3cm} | m{2.3cm} | }
    \hline
    & \multicolumn{2}{c|}{Identification} & \multicolumn{2}{c|}{Comparison} \\
    \hline
    \multirow{6}{2cm}{Condition \textit{low} $n_{obj} \in [3..8]$}
    & IDS\_3 & IDS\_3\_test & CDS\_3\_8\_0 & CDS\_3\_8\_test \\
    & IDS\_4 & IDS\_4\_test & CDS\_3\_8\_1 &  \\
    & IDS\_5 & IDS\_5\_test & CDS\_3\_8\_2 &  \\
    & IDS\_6 & IDS\_6\_test & CDS\_3\_8\_3 &  \\
    & IDS\_7 & IDS\_7\_test & CDS\_3\_8\_4 &  \\
    & IDS\_8 & IDS\_8\_test &  &  \\
    \hline
    \multirow{12}{2cm}{Condition \textit{mid} $n_{obj} \in [9..20]$}
    & IDS\_9  & IDS\_9\_test & CDS\_9\_20\_0 & CDS\_9\_20\_test \\
    & IDS\_10 & IDS\_10\_test & CDS\_9\_20\_1 &  \\
    & IDS\_11 & IDS\_11\_test & CDS\_9\_20\_2 &  \\
    & IDS\_12 & IDS\_12\_test & CDS\_9\_20\_3 &  \\
    & IDS\_13 & IDS\_13\_test & CDS\_9\_20\_4 &  \\
    & IDS\_14 & IDS\_14\_test &  &  \\
    & IDS\_15 & IDS\_15\_test &  &  \\
    & IDS\_16 & IDS\_16\_test &  &  \\
    & IDS\_17 & IDS\_17\_test &  &  \\
    & IDS\_18 & IDS\_18\_test &  &  \\
    & IDS\_19 & IDS\_19\_test &  &  \\
    & IDS\_20 & IDS\_20\_test &  &  \\
    \hline
    \multirow{9}{2cm}{Condition \textit{high} $n_{obj} \in [21..30]$}
    & IDS\_21 & IDS\_21\_test & CDS\_21\_30\_0 & CDS\_21\_30\_test \\
    & IDS\_22 & IDS\_22\_test & CDS\_21\_30\_1 &  \\
    & IDS\_23 & IDS\_23\_test & CDS\_21\_30\_2 &  \\
    & IDS\_24 & IDS\_24\_test & CDS\_21\_30\_3 &  \\
    & IDS\_25 & IDS\_25\_test & CDS\_21\_30\_4 &  \\
    & IDS\_26 & IDS\_26\_test &  &  \\
    & IDS\_27 & IDS\_27\_test &  &  \\
    & IDS\_28 & IDS\_28\_test &  &  \\
    & IDS\_29 & IDS\_29\_test &  &  \\
    & IDS\_30 & IDS\_30\_test &  &  \\
    \hline
\end{tabular}
\label{summary_table}
\vskip 0.5cm
\caption{Summary Table for SpatialSim, listing all datasets. The two main columns correspond to the two tasks. The three main rows correspond to the three object number condition: low, mid, and high. For each task/object number condition combination, the different datasets are listed according to whether they are train or test datasets. Validation datasets are omitted from the table for clarity, but are drawn from the same distribution as the test sets, and are available at the provided link. Note that Identification has a dataset for each configuration (one per number of objects) and that Comparison has five train dataset for each valid/test set corresponding to the curriculum in rotation angles described above.}
\end{center}
\end{table}

As described in the main text, the Comparison task is harder to train on than the Identification task. This is because of the presence of rotations in the allowed transformation for the same similarity class. This problem does not show when rotations are not included in the dataset. To help the optimization process, we generate a curriculum of datasets with a set of increasing ranges for allowed rotation angles $\theta$, up to the entire $[0, 2\pi]$ range. We thus generate, for each $n_{obj}$ condition (\textit{low}, \textit{mid}, \textit{high}) a set of 5 datasets with respective7
 allowed rotation angles $\theta$:

\begin{itemize}
    \item $\theta \in [0, \frac{\pi}{10}]$
    \item $\theta \in [0, \frac{\pi}{2} + \frac{\pi}{10}]$
    \item $\theta \in [0, \pi + \frac{\pi}{10}]$
    \item $\theta \in [0, \frac{3\pi}{2} + \frac{\pi}{10}]$
    \item $\theta \in [0, 2\pi]$
\end{itemize}

For each condition the test set is unique and has $\theta \in [0, 2\pi]$: we test on unrestrained rotations. This curriculum is used with all our models in all our experiments.

\textbf{Names of the datasets:} the datasets presented in Table \ref{summary_table} are named in the following way. 

\begin{itemize}
    \item \textbf{For Identification}, the 'IDS' prefix is followed by $n_{obj}$ and then by the '\_valid' and '\_test' suffix respectively for validation and test sets.
    \item \textbf{For Comparison}, the 'CDS' prefix is followed by the range of numbers of objects (the dataset may contain samples with $n_{obj}$ in this range, inclusive). The training datasets additionaly have an identifier corresponding to their place in the rotation angle curriculum (0 to 4, in the above-defined order). The validation and test have the '\_valid' and '\_test' suffix, respectively.
\end{itemize}

\section{Additional details on Dataset Generation} \label{section:dscreation}

In this section we give additional information on dataset creation. We consider the world as square with length and width $20$ units. We sample the x and y positions of our objects in this square. The sizes of our objects describe their radius (an object of size $s$ is contained in a square of side $2s$) and range from $0.5$ to $2$ units. For orientation, we used the following approximation: we considered orientation as a one-dimensional variable, expressed in radians, and we sample the objects' orientation between $0$ and $2\pi$. This is an approximation because the periodic nature of angles cannot be represented in one dimension. The colors of the objects are sampled in the continuous 3d RGB space, and each component ranges from $0$ to $1$. As for shapes, there are 3 possible categories (square, circle, triangle) that are represented by a corresponding one-hot vector.

\section{Models and Architectures} \label{section:models}

\subsection{Models for Identification}

\begin{figure*}[!htbp]
\vskip 0.2in
\begin{center}
\centerline{\includegraphics[width=420pt]{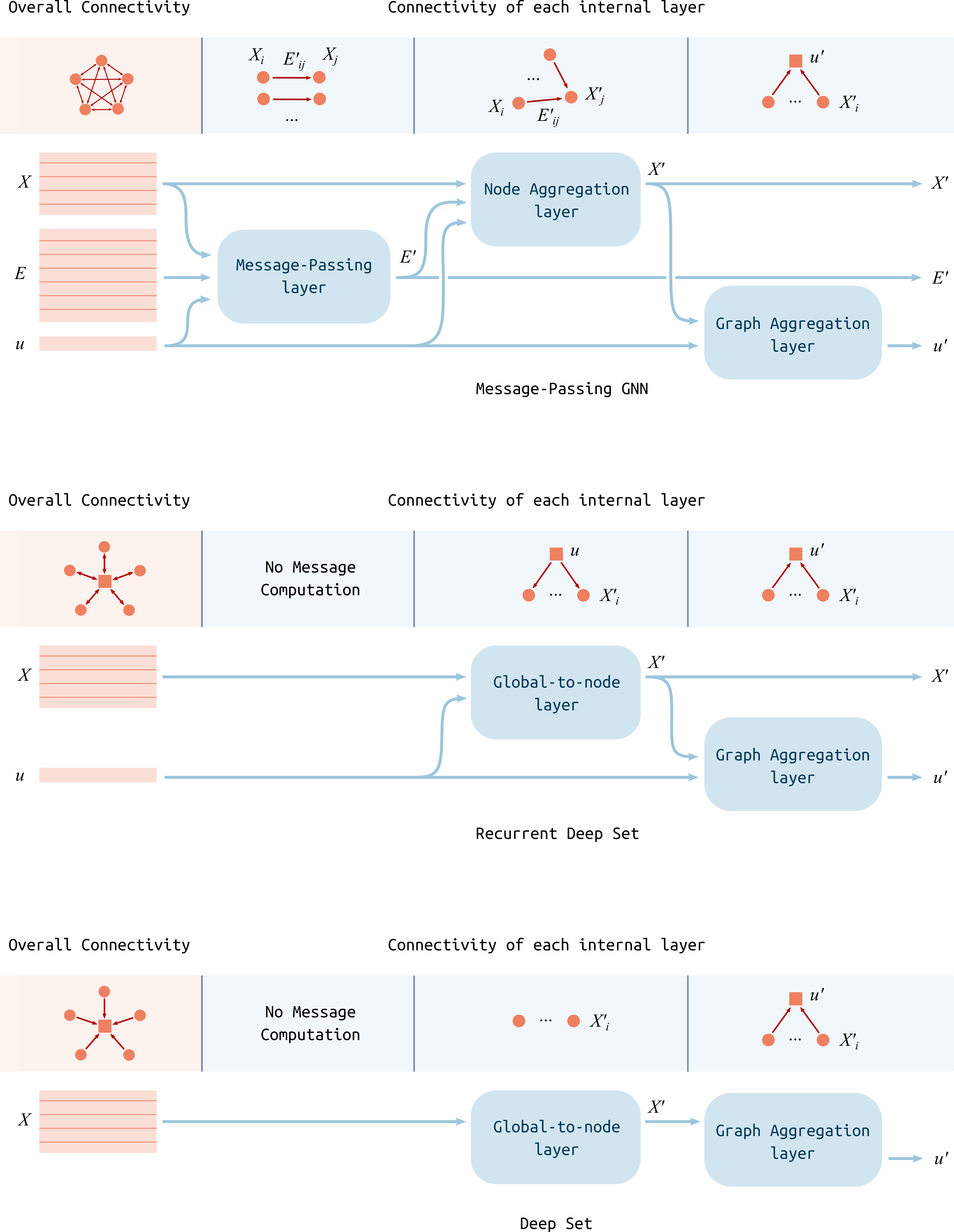}}
\caption{An illustration of the three different layers used in this work. Going from MPGNN to RDS to DS can be seen as an ablation study, where different elements are withdrawn from the layer to study their impact on final performance. For the MPGNN and RDS layers, the output tensors are then fed back as inputs of the model, providing recurrent computation; this is not the case for the Deep Set layer. In this figure, emphasis is put on the connectivity implied by each layer. Nodes are represented by orange disks, the graph-level embedding, which can be seen as a special kind of node, is represented with an orange square. From top to bottom, we go from all-to-all connectivity to bidirectional all-to-one to unidirectional all-to-one.}
\label{layers}
\end{center}
\vskip -0.2in
\end{figure*}

In this section we present our graph creation procedure for the Identification task and provide the equations for the models we use: Message-Passing GNN, Recurrent Deep Set, Deep Set and MLP. We additionally present a visual illustration of our different layers in Figure \ref{layers}.

\subsubsection{Graph Creation}

From a set of objects $S$ we construct a fully-connected, directed graph $G$ that is used as an input to our GNN. In our work, $G = (X, A, E, u)$ contains the following information :

\begin{itemize}
    \item $X \in \mathbb{R}^{n \times d_x}$ is a tensor of node features, containing a vector of dimension $d_x$ for each of the objects in the scene;
    \item $A \in \mathcal{M}^{n \times n}$ is the adjacency matrix of the graph;
    \item $E \in \mathbb{R}^{e \times d_e}$ is a tensor of edge features, also referred to as messages in the rest of this article, labeling each of the $e$ edges with a $d_e$-dimensional vector, and that can be seen as information propagating from the sender node to the receiver node. We choose the dimensionality of edges to be twice the dimensionality of nodes $d_x$;
    \item $u$ is a graph-level feature vector, used in the GNN computation to store information pertaining to the whole graph, and effectively used as an embedding of the graph to predict the class of the input.
\end{itemize}

\textbf{Initialization of the graph} : Since our models require inputs for $E$ and $u$ that are not \textit{a priori} given in the description of the collection of objects, we use a generic initialization scheme : $u$ is initialized with the mean of all node features, and each edge is initialized with the concatenation of the features of the sender node and the receiver node.

\subsubsection{Message-Passing GNN}

The MPGNN can be seen as a function operating on graph input and producing a graph output: $GNN : G(X, A, E, u) \rightarrow G'(X', A, E', u')$, where the dimensionnality of the node features, edge features and global features can be changed by the application of this function, but the graph structure itself encoded as the adjacency matrix $A$ is left unchanged. This GNN can then be described as the composition of several functions, each updating a part of the information contained in the graph :

\textbf{Message computation} : We denote by $E_{i \rightarrow j}$ the feature vector of the edge departing from node $i$ and arriving at node $j$, $X_i$ the feature vector of node $i$, and $[x || y]$ the concatenation of vectors $x$ and $y$, and by $MLP$ a multi-layer perceptron. The message passing step is then defined as : 
$$E'_{i \rightarrow j} \leftarrow MLP_E([X_i || X_j || E_{i \rightarrow j} || u])$$

At each time step, the message depends on the features of the sender and receiver nodes, the previous message, and the global vector $u$.

\textbf{Node-wise aggregation} : Once the message along each edge is computed, the model computes the new node features from all the incoming edges. We define by $\mathcal{N}(j)$ the incoming neighbourhood of node $j$, that is, the set of nodes $i \in \mathbb{[} 1.. n \mathbb{]}$ where there exists an edge going from $i$ to $j$. The node computation is then performed as so :
$$X'_j \leftarrow MLP_X\bigg{(}\Big{[}X_j \Big{|}\Big{|} \sum_{i \in \mathcal{N}(j)} E'_{i \rightarrow j} \Big{|}\Big{|} u\Big{]}\bigg{)}$$

\textbf{Graph-wise aggregation} Finally, we update the graph-level feature, that we use as an embedding for classification, and that conditions the first and second time step of computation :
$$u' \leftarrow MLP_u\Big{(}\Big{[}\sum_{i} X'_i \big{|}\big{|} u\Big{]}\Big{)}$$

\textbf{Prediction} : the final step is passing the resulting vector $u$ through a final multi-layer perceptron to produce logits for our binary classification problem :
$$out \leftarrow MLP_{out}(u')$$
We use the same dimensionality for the output vectors as for the input vectors of the message computation, node aggregation and graph aggregation, and this allows us to stack $N$ GNN computations in a recurrent fashion.

\subsubsection{Recurrent Deep Sets}

We introduce a simpler model we term Recurrent Deep Sets (RDS). This model is introduced to provide a comparison point to the MPGNN and assess how useful relational inductive biases are in performing well on the benchmark. This method dispenses with the message computation and node aggregation part, and at each step only transforms the node features and aggregates them into the graph feature. This architecture is resembles the Deep Set, to the important difference that the graph-level feature $u$ is then fed back at the following step by being concatenated to the object feature for the next round of computation. This allows the computation of features for each object to depend on the state of the whole configuration, as summarized in the graph embedding $u$. This contrasts with the original Deep Sets, where each object is processed independently. The functional description of this model is thus :
$$X'_j \leftarrow MLP_X([X_j || u])$$
$$u' \leftarrow MLP_u\Big{(}\Big{[}\sum_{i} X'_i \big{|}\big{|} u\Big{]}\Big{)}$$
$$out \leftarrow MLP_{out}(u')$$
Note that for this model, there is no need to connect each object to every other object. However, this back-and-forth between node computation and graph aggregation can be interpreted as computing messages between each object and a central node, that represents the information of the whole graph. In this sense, this model can be interpreted as a GNN operating on the star-shaped graph of the union of the set of objects and the central graph-level node. In particular, this means that the resulting model performs a number of computations that scales linearly in the number of nodes, instead of quadratically as is the case for a message-passing GNN on the complete, fully connected graph of objects. While this is an interesting propriety, in practice for a fixed size of $u$ the number of objects cannot grow arbitrarily large because the success of our models depend on the ability of $u$ to accurately summarize information which is dependent on all the objects, which becomes difficult as the number $n$ of objects becomes large.

\subsubsection{Deep Sets}

In this section we summarize shortly the computations done by the Deep Set model. The model can be described as a node-wise transformation composed with a sum operator on all the nodes, followed by a final transformation. Namely, the Deep Set defines the following transformations:

$$X'_j \leftarrow MLP_X(X_j)$$
$$u' \leftarrow \Big{[}\sum_{i} X'_i \Big{]}$$
$$out \leftarrow MLP_{out}(u')$$

Note that, contrary to the MPGNN and the RDS, the Deep Set has no recurrent structure; running it several times will always produce the same output.

\subsubsection{Hyperparameters}

In our experimental setup, for a MPGNN/RDS/Deep Set we let $h$ be the dimension of the hidden layers for all internal MLPs ($MLP_E$, $MLP_X$, $MLP_u$, and $MLP_{out}$, when each of these MLPs are defined, as appropriate). We let $d$ be the number of hidden layers. We then have, to keep a similar number of parameters between GNN models, $h = 16$ and $d = 1$ for MPGNN, $h = 16$ and $d = 2$ for RDS, and $h = 16$ and $d = 4$ in Deep Set. We use ReLU non-linearities in each MLP. We use (for MPGNN and RDS) $N = 1$ successive passes through the GNN, since increasing $N$ did not seem to affect the performance in a significative way.

We also define the MLP baseline as having $d = 2$ layers of $h = n_{obj} \times 16$ hidden units. This was done to provide the MLP with a roughly comparable number of units to the GNNs (since the latter models maintain a hidden representation of size $16$ for each node). The number of units here refer to the cumulative dimensions of the hidden vectors, the number of parameters to the number of scalar weights and biases. In particular, this design was adopted because the number of hidden units 

\subsection{Models for Comparison}

To tackle this task, we construct from one sample of two configurations two different graphs, one representing each set of objects, in the same way as in Identification. In this section we introduce a straightforward Dual-Input Model (hereby referred as DIM) that operates on input pairs of graphs. The internal GNNs used inside the DIM can be any one of MPGNN, RDS or Deep Set, and we will identify different dual-input models by their internal component type.

\subsubsection{Dual-input architecture}

\begin{figure*}[!htbp]
\vskip 0.2in
\begin{center}
\centerline{\includegraphics[width=220pt]{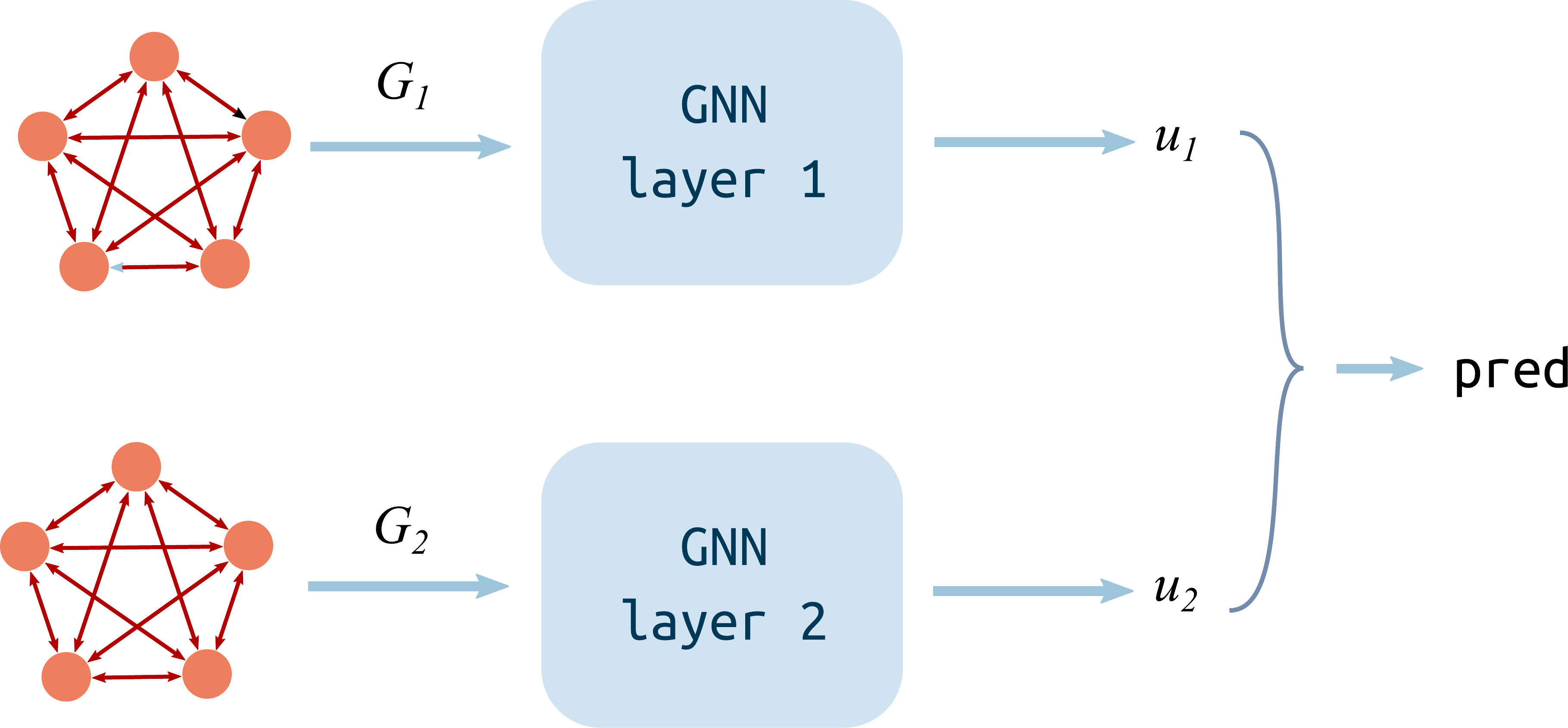}}
\caption{An illustration of the two dual-input architecture. Two parallel layers (MPGNN, RDS or Deep Set) process the input graphs in parallel, and the resulting global vectors are concatenated and passed through a final MLP.}
\label{archis}
\end{center}
\vskip -0.2in
\end{figure*}

Let us denote by $GNN$ a GNN layer, as defined in the discussion of Comparison architectures. The DIM is composed of two parallel GNN layers, $GNN_1$ and $GNN_2$. Each input graph is processed by its corresponding layer, as such:

$$X'_1, E'_1, u'_1 \leftarrow GNN_1(X_1, A_1, E_2, u_1)$$
$$X'_2, E'_2, u'_2 \leftarrow GNN_2(X_2, A_2, E_2, u_2)$$

As previously, we repeat this operation $N$ times, and we produce the output as:

$$out \leftarrow MLP_{out}([u'_1 || u'_2])$$

\subsubsection{Hyperparameters}

We use the same hyperparameters in for this task as in the previous one. The MLP baseline is also defined in the same way, except that the number of hidden units in each layer is doubled to account for the doubling in number of objects. Since the datasets used in Comparison contain a variable number of objects across samples, we use the mean $n_{obj}$ for determining the number of hidden units in the MLP.

\section{Model Heatmaps} \label{section:mhms}

This section provides additional discussion on the model heatmap visualizations presented in the eponymous section in the main text. We present more fully the description of what these visualizations mean and we provide additional commentary on the qualitative differences between models, conditions (\textit{low} number of objects, \textit{mid} number of objects and \textit{high} number of objects).

\subsection{Additional details on heatmap generation} 

Each one of the models we use in this work projects the input graph $G = (X, A, E, u)$ on a two-dimensional vector with coordinates $(\mathcal{C}_+, \mathcal{C}_-) \in \mathbb{R}^2$. These values correspond respectively to the scores (logits) for the positive and the negative classes: if $\mathcal{C}_+ \geq \mathcal{C}_-$ the input is classified as positive, otherwise it is classified as negative. To produce one heatmap image for an object of index $o_i$ of feature vector $X_i$, we plot $H = \mathcal{C}_+ - \mathcal{C}_-$ as a function of $o_i$'s x-y position, while holding $o_i$'s non-spatial features as well as all other object features constant. Thus, every pixel where $H$ is positive corresponds to an input with an alternative x-y position for $o_i$ that the model classifies as positive. The same thing holds for negative values of $H$: they correspond to positions of $o_i$ that would result in the input being classified as a negative. The actual prediction of the model for the given input is given by the color of the current position of $o_i$, marked by a star in our plots.

In this section we plot the heatmaps for Comparison models. We do this according to the previous description, by comparing a configuration with a copy of itself, and by moving an object in the copy configuration only; in this case $o_i$ refers to one of the objects in the copied configuration.

\subsection{Discussion}

\begin{figure}[!t]
    \centering
    \includegraphics[width=400pt]{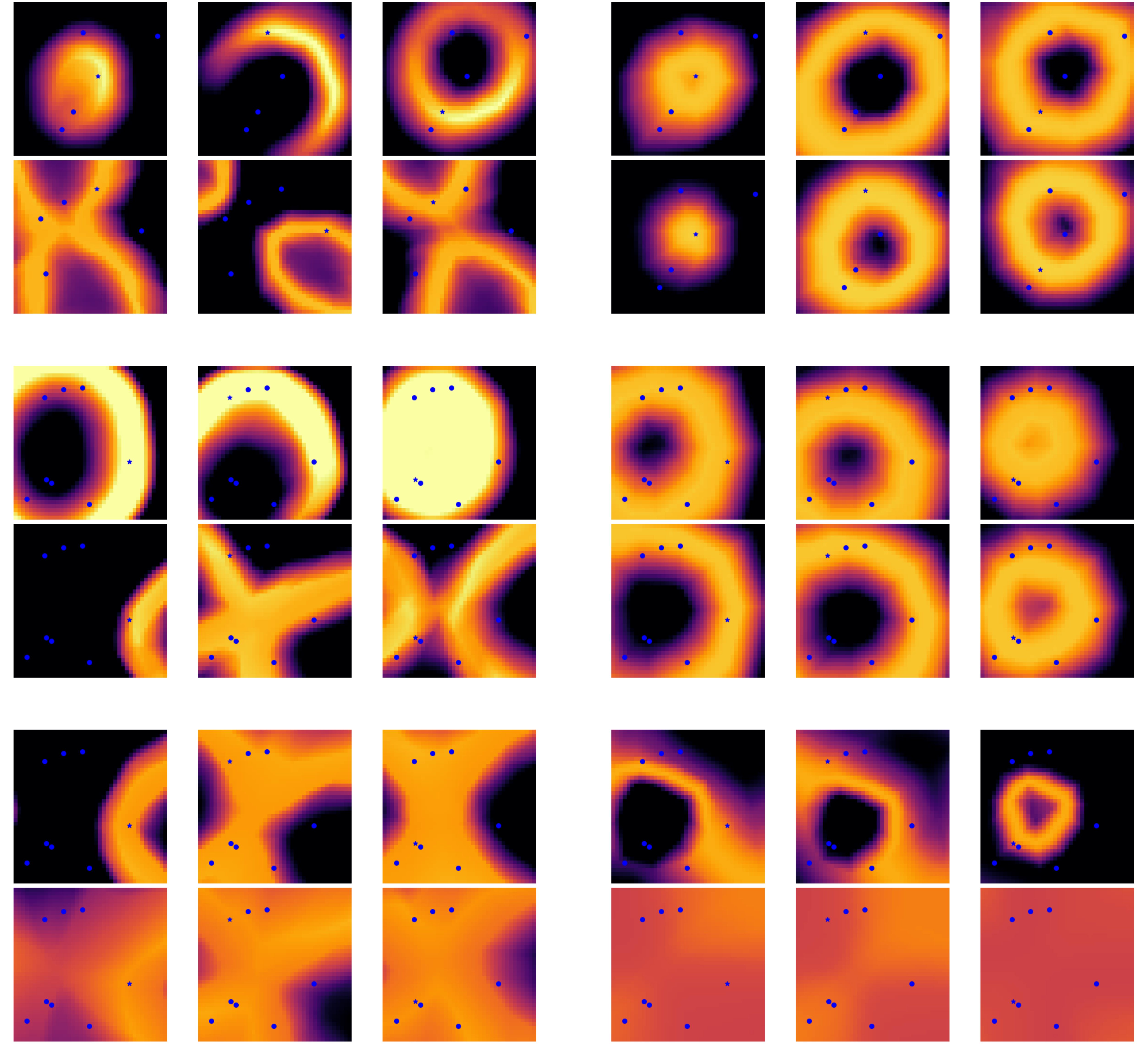}
    \caption{Model heatmaps for Comparison models. The plots are organized as follows: the left column corresponds to dual-input models with MPGNN internal layers, the right one plots dual-input models with RDS layers. Each of the larger-scale rows plots, respectively: models trained on \textit{low} numbers of objects ($n_{obj} \in [3..8]$) and 5 objects plotted, models trained on \textit{low} numbers of objects and plotted with 8 objects, and models trained on \textit{mid} numbers of objects ($n_{obj} \in [9..20]$) and plotted with 8 objects, for contrast. Within each of the six blocks, each three-image row corresponds to the heatmaps generated on one random training run of a model, and each image corresponds to moving about one particular object $o_i$. For each image, the fixed objects are represented by a blue dot corresponding to their position, and the perturbed object is identified with a blue star.}
    \label{fig:mhmdtop}
\end{figure}

\begin{figure}[!t]
    \centering
    \includegraphics[width=400pt]{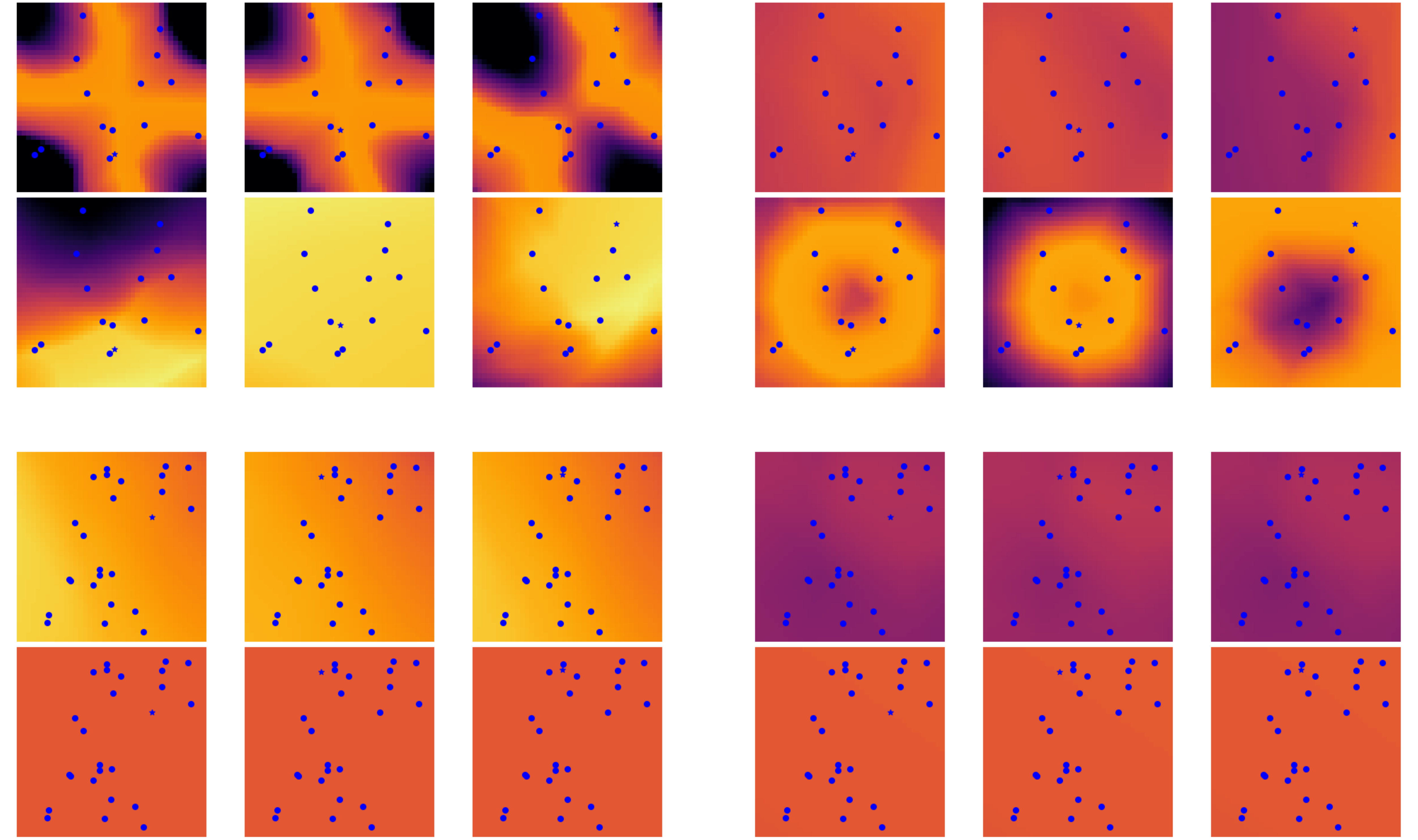}
    \caption{Model heatmaps for Comparison models. The left and right columns are respectively MPGNN and RDS, as in Figure \ref{fig:mhmdtop}. The large-scale rows correspond to models trained on \textit{mid} numbers of objects and plotted with a configuration of 15 objects, and models trained with \textit{high} ($n_{obj} \in [21..30]$) numbers of objects and plotted with 25 objects.}
    \label{fig:mhmdbot}
\end{figure}

The heatmaps are given in Figure \ref{fig:mhmdtop} and Figure \ref{fig:mhmdbot} for different models, training datasets, numbers of objects and seeds. Looking at these model heatmaps allows us to have a qualitative grasp of the functions learned by our different models, and in particular how well these functions encode the similarity classes they are trained to represent. The Deep Set models were not included in the figures because these models predict the same value of $H$ for each position of $o_i$. This means that, when holding the objects $o_j, j \neq i$, fixed, the model is (almost) invariant to changes in position of $o_i$.

Before going further, one should note that these plots allow us to visualize the variation of the models' learned function only with respect to two variables among many, and the portion of the variation we visualize becomes smaller as $n_{obj}$ grows, because adding objects is adding variables. Nevertheless, these variations are important because they allow us to probe the boundaries of what our models classify as being the same configuration as opposed to what they classify as being different configurations.

One of the first thing we can note is the qualitative difference between MPGNN and RDS models, espescially when $n_{obj}$ is low (Figure \ref{fig:mhmdtop}). RDS heatmaps seem to consistently exhibit a ring-like structure, with the areas corresponding to the positive class form a ring centered around the center of the configuration and passing through $o_i$. We conclude from this that the model has leaned to use the distance from the center of the configuration (which has a good chance of being different for each object of a random configuration) as one of the main features in classifying its input. This is to be expected when we look at the computations done by the RDS: each node has access to an average of all the other nodes before performing its own node update. MPGNNs sometimes learn ringlike structures that seem more modulated as in the case of RDS, sometimes being open rings. Other times, MPGNNs heatmaps exhibit a kind of cross-like structure, or two symmetrical rings; $o_i$ is placed at one of the high-value spots of this structure (indicating that the model has learned to assign the positive class to a set of two identical copies of the same configuration). These structure seem to exhibit symmetry with respect to the principal axis of the configuration, suggesting that MPGNN learns to compute and use this as a feature when tasked to compare two different configurations it never has seen before. The different forms of the trained MPGNNs may also hint at a higher expressivity of the model, its ability to approximate a wider range of functions.

Another interesting thing this visualization allows us to see is the difference in functions learned by models on two different datasets. Figure \ref{fig:mhmdtop}'s second and third rows compare models on the same configuration of 8 objects, but the ones in the second row have been trained with $n_{obj} \in [3..8]$ whereas the ones in the third row have been trained with $n_{obj} \in [9..20]$. The function learned exhibit qualitative differences, even if the presented configuration and the models are he same, as a result of different training conditions. The heatmaps in the bottom row appear more spread out. We take it to show that the functions learned while training on higher numbers of objects are less sensitive to the variation of a single object's position. This is probably so because of the way the negative samples are created in our datasets: randomly resample all object positions (and then rotate, scale, and translate all objects randomly). As $n_{obj}$ gets larger, the compared examples presented have a very high probability to be widly different from the target configuration, making the model less likely to learn about the contribution of the perturbation of only one object.

Figure \ref{fig:mhmdbot} corroborates this view: the functions learned exhibit much less variation to the perturbation of the position of a single object, particularly in the \textit{high} ($n_{obj} = 25$) case. This figure also showcases a prediction error: the top row of the bottom-right block is a visualization of an RDS model that assigns the negative class to all alternative positions of the object $o_i$, including its current one. This should not surprise us: when training with a \textit{high} number of objects, many RDS models do not train and perform only slightly above chance.

\section{Easier and Harder Configurations to Identify} \label{section:easyhard}

\begin{figure}[!htbp]
\vskip 0.2in
\begin{center}
\centerline{\includegraphics[width=400pt]{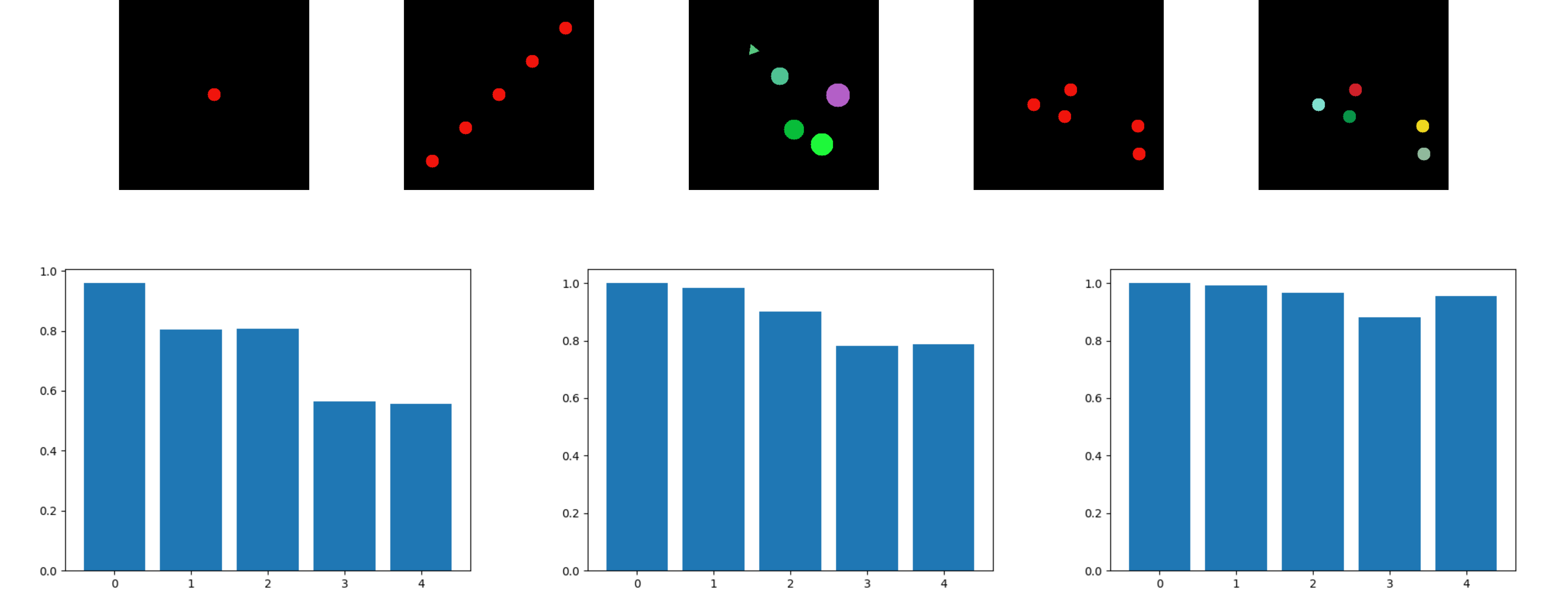}}
\caption{The top row represents the configurations we trained our models with, as described in the text. The bottom row is a bar plot of the final test accuracy of (from left to right) the Deep Set, Recurrent Deep Set and Message-Passing GNN on each of the 5 datasets, in the order specified in the top row (results were computed on 5 seeds for each dataset).}
\label{easyhard}
\end{center}
\vskip -0.3in
\end{figure}

In this section we study why particular configurations may be harder or easier to recognize for our models, in the context of the Identification task. We hypothesise that more regular arrangements of objects must be easier to tell apart than more random configurations, and that configurations with a high degree of object diversity (many colors, many shapes) must also be easier to learn to classify, because the models can more easily identify and match the different objects. To test this, we compare one randomly generated dataset (regular difficulty) with 1) a configuration where all objects are red circles of the same size positioned at the same point; 2) a configuration where all the objects are red circles of the same size arranged in a line; 3) a configuration where all the objects are randomly positioned red circles of the same size; and 4) the same configuration as 3), but with circles of varying color. We train our three layers, DS, RDS and MPGNN, to recognize these configurations, and report the results in Figure \ref{easyhard}, along with an illustration of the configurations.

We interpret the results as follows : the fourth configuration, the one with all red circles, does seem to be more difficult to learn across all models. This may be due to the intrinsic hardness of the task on this configuration, or to the fact that randomly resampled positions for the negative examples of this dataset may give with non-negligible probability configurations that are close to a translated/rotated version of the reference example, because any object can be identified with any other. This second option may translate into negative examples that may resemble strongly positive examples, confusing the model. Interestingly, the problem fades when we identify each object by giving it a color, suggesting this second explanation is correct, but only for the MPGNN. For MPGNN, performance is roughly similar on all the other datasets. However, for DS and RDS, there seems to be considerable difference between datasets. The DS layer fails to perform significantly above chance for both right-hand configurations, suggesting arrangements of similar objects are difficult for this kind of model. Interestingly, the DS layer performs similarly on the aligned red circles than on the random diverse configuration, but significantly better than on the configuration with randomly scattered red circles, suggesting it is able to use the alignment information to reach above-chance accuracy, but not in a completely reliable way. As a contrast, the RDS layer performs near-perfectly on this configuration, showing that the additional connectivity of the RDS helps it in discovering exploitable regularities in the data.

\section{Generalization to Other Number of objects} \label{section:gen}

In this section we present some generalization results for the Comparison task. Since the models for this task are trained on any couple of configurations, they can be transferred to datasets with higher numbers of objects. In this experiment we train Deep Set, RDS and MNGNN models on one dataset ($n_{obj} \in [3..8]$, $n_{obj} \in [9..20]$ or $n_{obj} \in [21..30]$) and test the models on all three datasets. The results are reported in Table \ref{genres}.

\begin{table}[!h]
\begin{center}
\begin{tabular}{ | m{1.3cm} || m{2.3cm}| m{2.3cm} | m{2.3cm} | }
    \hline
    & 3-8 & 9-20 & 21-30 \\
    \hline
    \hline
    \multirow{3}{2cm}{3-8}
    & 0.51 $\pm$ 0.016 & 0.49 $\pm$ 0.046 & 0.50 $\pm$ 0.043 \\
    & 0.80 $\pm$ 0.133 & 0.66 $\pm$ 0.138 & 0.51 $\pm$ 0.048 \\
    & \textbf{0.89} $\pm$ 0.03 & \textbf{0.71} $\pm$ 0.092 & \textbf{0.56} $\pm$ 0.075 \\
    \hline
    \multirow{3}{2cm}{9-20}
    & 0.51 $\pm$ 0.046 & 0.50 $\pm$ 0.001 & 0.50 $\pm$ 0.047 \\
    & \textbf{0.75} $\pm$ 0.125 & 0.68 $\pm$ 0.154 & 0.52 $\pm$ 0.054 \\
    & 0.68 $\pm$ 0.063 & \textbf{0.81} $\pm$ 0.121 & \textbf{0.68} $\pm$ 0.16  \\
    \hline
    \multirow{3}{2cm}{21-30}
    & 0.50 $\pm$ 0.04 & 0.51 $\pm$ 0.068 & 0.50 $\pm$ 0.05 \\
    & \textbf{0.60} $\pm$ 0.087 & 0.68 $\pm$ 0.15 & 0.52 $\pm$ 0.04  \\
    & 0.51 $\pm$ 0.048 & \textbf{0.77} $\pm$ 0.12 & \textbf{0.71} $\pm$ 0.18  \\
    \hline
\end{tabular}
\label{genres}
\vskip 0.5cm
\caption{Generalization results between datasets for Deep Set, RDS and MPGNN. The numbers plotted are averages of testing accuracies. Columns correspond to training datasets, rows to testing datasets. Each block corresponds to one train-set/test-set combination. In each block, the results are given from top to bottom for Deep Set, RDS and MPGNN. Diagonal blocks correspond to matching train set/test set combinations. All reported results are averages and standard deviations over 10 different runs. Rows and columns are annotated with the $n_{obj}$ range.}
\end{center}
\end{table}

The results demonstrate the limited abilities of the models to transfer their learned functions to higher or lower numbers of objects. For instance, MPGNNs achieve 0.89 test accuracy when trained \textit{and} tested on 3 to 8 objects, but this performance decreases sharply on the datasets with higher numbers of objects. This is less the case for RDS, presumably because the simpler functions they learn, while achieving lower performance when tested on the matching dataset, are more robust to higher numbers of objects. Another interesting point is that models trained on 9 to 20 numbers of objects appear to transfer better than other conditions. In particular, both RDS and MPGNN achieve higher mean test accuracy when transferring from 9-20 objects to 21-30 objects than models which were directly trained on 21-30 numbers of objects. The 21-30 dataset is harder to train on, so the models trained directly on this dataset may never learn, which bring the mean accuracy down. This suggests that functions useful for good performance on 9-20 numbers of objects are also useful for 21-30 numbers of objects. In contrast, functions useful for good performance on 3-8 numbers of objects do not transfer well to higher numbers of objects. 

These suggest a tradeoff in being able to solve the task well for low numbers of objects versus being able to solve the task for high numbers of objects. This confirms the qualitative evaluation in Section \ref{section:mhms}, where we remarked that the functions learned by the models varied greatly with the dataset they were trained on.

\section{Training on Less Examples} \label{section:trainvar}

In this section we vary the number of unique examples presented to the models in the training set. We keep the same number of optimizer steps as in the main experiments, but we reduce the number of samples we train on. The results for Identification are presented in Table 3, and the results for Comparison are reported in Table 4.

In both Tables, in the first two rows we see all models overfitting the dataset, their test accuracy being at 0.5. They are unable to transfer to the training set and performing at chance levels. Then, respectively at 1000 samples for Identification and at 10000 samples for Comparison the performance levels rise very close to their final levels. We wanted to observe whether the additional relational inductive biases in MPGNNs would allow for faster training than RDS and Deep Set; however we do not observe this: all models seem to have similar progression levels as the size of the training set increases. From this we conclude that the advantage of MPGNNs do not stem from their sample-efficiency, but rather from their ability to represent more complex functions.

\begin{table}[!h]
\begin{center}
\begin{tabular}{lccr}
\toprule
& MPGNN & RDS & Deep Set \\
\midrule
10 & 0.52 $\pm$ 0.038 & 0.52 $\pm$ 0.035 & 0.52 $\pm$ 0.032 \\
100 & \textbf{0.64} $\pm$ 0.051 & 0.58 $\pm$ 0.035 & 0.54 $\pm$ 0.019 \\
1000 & \textbf{0.94} $\pm$ 0.041 & 0.86 $\pm$ 0.065 & 0.61 $\pm$ 0.036 \\
10000 & \textbf{0.97} $\pm$ 0.026 & 0.91 $\pm$ 0.062 & 0.65 $\pm$ 0.079 \\
\bottomrule
\end{tabular}
\label{table3}
\vskip 0.5cm
\caption{Mean accuracies for training on reduced numbers of examples on Identification. The last row represents the full training set.}
\end{center}
\end{table}

\begin{table}[!h]
\begin{center}
\begin{tabular}{lccr}
\toprule
& MPGNN & RDS & Deep Set \\
\midrule
100 & 0.50 $\pm$ 0.005 & 0.50 $\pm$ 0.004 & 0.50 $\pm$ 0.005 \\
1000 & 0.50 $\pm$ 0.004 & 0.50 $\pm$ 0.003 & 0.50 $\pm$ 0.005 \\
10k & \textbf{0.87} $\pm$ 0.016 & 0.82 $\pm$ 0.098 & 0.52 $\pm$ 0.01 \\
100k & \textbf{0.89} $\pm$ 0.03 & 0.80 $\pm$ 0.133 & 0.51 $\pm$ 0.014 \\
\bottomrule
\end{tabular}
\label{table4}
\vskip 0.5cm
\caption{Mean accuracies for training on reduced numbers of examples on Comparison. The last row represents the full training set.}
\end{center}
\end{table}

\section{Adding Distractor Objects} \label{section:distractors}

In realistic environments cluttered with objects, only some of the objects could be relevant for the similarity task at hand; some of the objects may be distractors unrelated to the task. To test how robust our models are to additional objects in the input that bear no relevance to the task, we generate additional train and test sets for $n_{obj} \in [3..8]$. We use numbers of distractors $n_d \in [0..3]$ for both Identification and Comparison. The results are reported in Table \ref{table:distractors}.

\begin{table}[!h]
\caption{Test accuracies on the distractor datasets.}
\label{table:distractors}
\vskip 0.15in
\begin{center}
\begin{small}
\begin{sc}
\begin{tabular}{lcr}
\toprule
Model    & Identificatiom & Comparison \\
\midrule
MPGNN         & \textbf{0.87} $\pm$ 0.043 & \textbf{0.76} $\pm$ 0.019 \\
RDS           & 0.78 $\pm$ 0.102  & 0.59 $\pm$ 0.069  \\
Deep Set      & 0.67 $\pm$ 0.073  & 0.51 $\pm$ 0.01 \\
\bottomrule
\end{tabular}
\end{sc}
\end{small}
\end{center}
\vskip -0.1in
\end{table}

We see the model performance consistently drop for MPGNN and RDS, with a decrease in test accuracy of around 10\% on both tasks. The distractors seem to have no effect on Deep Set performance, suggesting that Deep Sets do not rely on a precise representation of object configuration. Dealing effectively with distractor objects could be done by adding n attention mechanism to the GNNs, a topic we leave for further work.